%% file: bare_jrnl_new_sample4.tex
\documentclass[lettersize,journal]{IEEEtran}
\usepackage{amsmath,amsfonts}
\usepackage{algorithmic}
\usepackage{algorithm}
\usepackage{array}
\usepackage[caption=false,font=normalsize,labelfont=sf,textfont=sf]{subfig}
\usepackage{textcomp}
\usepackage{stfloats}
\usepackage{url}
\usepackage{verbatim}
\usepackage{graphicx}
\usepackage{cite}
\usepackage{multirow}
\usepackage{enumitem}
\begin{document}

\title{From a Social Cognitive Perspective: Context-aware Visual Social Relationship Recognition}

\author{Shiwei Wu, Chao Zhang, Joya Chen, Tong Xu,~\IEEEmembership{Member,~IEEE}, Likang Wu, Yao Hu, \\ Enhong Chen, ~\IEEEmembership{Fellow,~IEEE}
\thanks{
\IEEEcompsocthanksitem Shiwei Wu, Chao Zhang, Tong Xu, Likang Wu, and Enhong Chen are with the Anhui Province Key Lab of Big Data Analysis and Application, School of Data Science and School of Computer Science, University of Science and Technology of China. E-mail: \{dwustc, wulk\}@mail.ustc.edu.cn, zclfe00@gmail.com, \{tongxu, cheneh\}@ustc.edu.cn
\IEEEcompsocthanksitem Joya Chen is with the College of Design and Engineering, National University of Singapore. E-mail: joyachen@u.nus.edu
\IEEEcompsocthanksitem Yao Hu is with the Xiaohongshu.Inc E-mail: xiahou@xiaohongshu.com}
}


\markboth{IEEE TRANSACTIONS ON PATTERN ANALYSIS AND MACHINE INTELLIGENCE,~Vol.~X, No.~X, X~2023}%
{Shell \MakeLowercase{\textit{et al.}}: A Sample Article Using IEEEtran.cls for IEEE Journals}


\maketitle

\input{chapters/00_abstract}

\begin{IEEEkeywords}
social relationship, multi-modal analysis, vision-linguistic contrasting.
\end{IEEEkeywords}

\input{chapters/01_intro}
\input{chapters/02_related}
\input{chapters/03_method}
\input{chapters/04_exp}
\input{chapters/05_conclusion}
\bibliographystyle{ieeetr}
\bibliography{reference}

\vfill

\end{document}

%% file: chapters/00_abstract.tex
People's social relationships are often manifested through their surroundings, with certain objects or interactions acting as symbols for specific relationships, e.g., wedding rings, roses, hugs, or holding hands.
This brings unique challenges to recognizing social relationships, requiring understanding and capturing the essence of these contexts from visual appearances. However, current methods of social relationship understanding rely on the basic classification paradigm of detected persons and objects, which fails to understand the comprehensive context and often overlooks decisive social factors, especially subtle visual cues. To highlight the social-aware context and intricate details, we propose a novel approach that recognizes \textbf{Con}textual \textbf{So}cial \textbf{R}elationships (\textbf{ConSoR}) from a social cognitive perspective. Specifically, to incorporate social-aware semantics, we build a lightweight adapter upon the frozen CLIP to learn social concepts via our novel multi-modal side adapter tuning mechanism. Further, we construct social-aware descriptive language prompts (e.g., scene, activity, objects, emotions) with social relationships for each image, and then compel ConSoR to concentrate more intensively on the decisive visual social factors via visual-linguistic contrasting. Impressively, ConSoR outperforms previous methods with a 12.2\% gain on the People-in-Social-Context (PISC) dataset and a 9.8\% increase on the People-in-Photo-Album (PIPA) benchmark. Furthermore, we observe that ConSoR excels at finding critical visual evidence to reveal social relationships.


%% file: chapters/01_intro.tex
\section{Introduction}

\begin{figure}[t]
  \includegraphics[width=0.5\textwidth, height=0.35\textwidth]{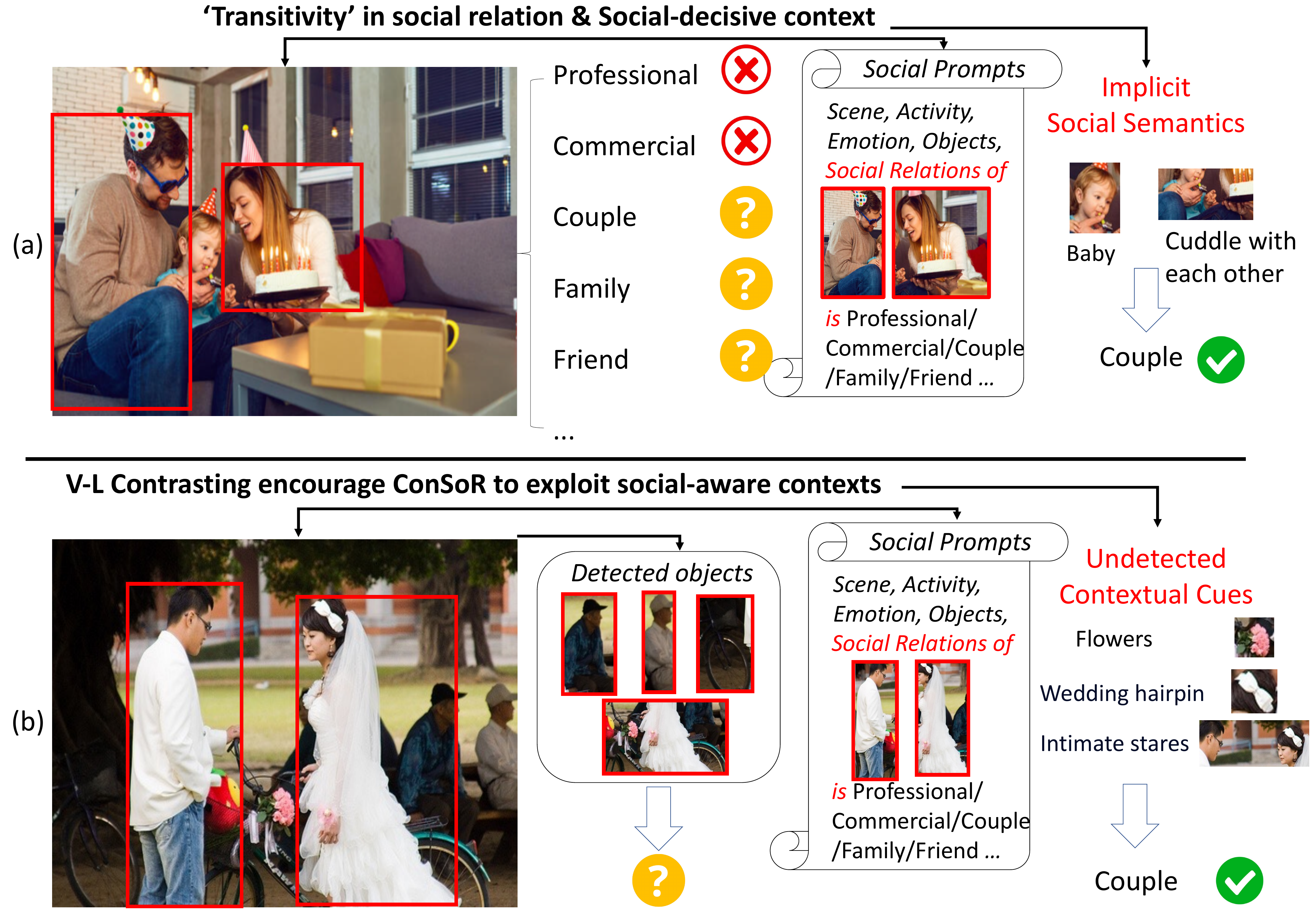}
  \caption{ConSoR excels at identifying decisive visual social cues. The explicit clues (e.g., \textit{party scene} in (a)) determine the probable intimate relation, while ConSoR aids in uncovering implicit social cues (e.g., the presence of the \textit{child} and their \textit{cuddle} interaction), further pinpointing the accurate \textit{Couple} relation. Additionally, the observed `transitivity' property in social relations also supports this conclusion. In (b), the visual-linguistic contrasting helps to capture the undetected yet critical contextual social cues~(e.g., \textit{flowers} and \textit{intimate stares}), which have been overlooked by previous methods.}
  \label{fig:intro}
\end{figure}

\begin{quote}
    ``No man is an island, \\
    entire of itself; \\
    every man is a piece of the continent, \\
    a part of the main." \\
    — John Donne, \textit{No Man Is an Island}
\end{quote}

In today's interconnected world, understanding \textit{social relationships} has grown increasingly vital across numerous domains. Gaining insights into social relations contributes beneficially to human health behaviors, encompassing chronic illness self-management and reducing suicidal tendencies~\cite{tay2013social, holt2010social, umberson2010social}; facilitates the design of intelligent robots with emergent ``social behavior''~\cite{de1960learning, beckers2000fom, deneubourg1991dynamics, fong2003survey}; and also underpins semantic-related services, including personalized content recommendations and targeted advertising~\cite{tang2016recommendation, khelloufi2020social, guo2023recommendation}. 

Unfortunately, while we humans can easily comprehend social relations among individuals through various comprehensive cues such as interactions, facial expressions, clothing styles, common patterns, and rich contextual information~\cite{heider2013psychology, fiske1992four}, it remains an extremely challenging task for intelligent systems to automatically capture social relations by comprehending these massive potential cues, primarily due to the huge gap between visual evidence and high-level semantics. This limitation severely restricts the application of social relation factors in various domains. To address this issue, numerous efforts have been made employing a wide range of technical approaches.
Most existing methods primarily focus on tackling this task through structural modeling of people and detected objects based on interactions or co-occurrences~\cite{li2017dual, wang_deep_2018, li2020graph, kukleva2020learning}, following a visual relation classification paradigm from a purely visual perspective. 
However, these previous methods cannot capture the critical yet implicit social clues effectively and often miss the decisive factor since they are constrained within the limited detected objects (e.g., within 80 categories in COCO-based~\cite{lin2014microsoft} detectors), and a lack of deep semantic knowledge about certain objects or interactions (e.g., a man and a woman, both wearing wedding rings, are holding hands, suggesting that they are a \textit{couple}, while `rings' are not even included in the COCO object categories).
To tackle this issue, we utilize a visual-linguistic contrasting approach, encouraging models to concentrate on the social-decisive visual factors, as well as incorporating social-aware semantics from pre-trained CLIP~\cite{radford2021learning} models, which further capture implicit social cues from the comprehensive context.

Let's examine Figure~\ref{fig:intro} as an example. While we can infer that the social relation between the man and woman shown in Fig.~\ref{fig:intro}a is likely intimate based on explicit semantics(e.g., the party scene), it is still challenging to distinguish the specific social relation among similar intimate relationships~(\textit{Couple, Family or Friend}). However, by inducing the model to concentrate on the social-aware contexts, ConSoR can capture the implicit yet critical social-relevant contextual cues, such as the presence of a child in the scene and their cuddling interaction. This helps to identify that the individuals are parents with children, further confirming that the two adults are \textit{couple}. This scenario also exemplifies the `transitivity' property in social relations. Moreover, the visual-linguistic contrasting with the descriptive social prompts approach helps ConSoR focus on the undetected yet decisive social clues~(\textit{flowers, intimate stares} in Figure~\ref{fig:intro}b), which have been ignored by the previous approaches.

Motivated by these observations, in this paper, we propose \textbf{ConSoR} to concentrate on the social-cognitive contexts through the visual-linguistic contrasting approach, further recognizing social relations among individuals.
Specifically, we first introduce the novel Multi-modal Side Adapter Tuning~(MSAT), which transfers rich semantic knowledge from both CLIP visual and text encoders into our lightweight multi-modal backbone through a hybrid feature fusion strategy. Moreover, in the visual branch, we perform relation reasoning from both context and interpersonal relation perspectives to examine the visual factors that help determine social relations among individuals, without the heavy object detector. 
Simultaneously, in the linguistic branch, we move beyond simple one-hot vector classification, which overlooks the rich information language offers. Instead, we utilize social-relevant contextual corpora to generate descriptive prompts for each category label. Notably, the candidate contextual social descriptive prompts for each image differ only in the social relation category. Therefore, employing a visual-linguistic contrasting approach allows the model to infer specific social relations more accurately, concentrate on crucial social factors, and comprehend intrinsic semantic connections among extensive social clues. This method also introduces a degree of inherent explainability. Our technical contributions are summarized as follows:

\begin{itemize}[leftmargin=*]

    \item  We propose ConSoR, which transfers rich semantics from CLIP via the novel Multi-modal Side Adapter Tuning mechanism, and further employs a visual-linguistic contrasting approach to steer models towards focusing on critical social-decisive contexts.

    \item Our innovative Contextual Interpersonal Reasoning module demonstrates a simple yet effective approach for relation reasoning. This design facilitates not only the modeling of the intrinsic social properties (e.g., transitivity, symmetry, and reflexivity), but also the exploitation of potential contextual cues without relying on a heavy object detector.

    \item We incorporate linguistic information through descriptive social prompts without extra annotations, fully utilizing the rich information language offers, beyond the limitations of traditional one-hot vector classification.
    
    \item Extensive validations on benchmark datasets demonstrate that ConSoR significantly outperforms SOTA baselines.
    
\end{itemize}

%% file: chapters/02_related.tex
\section{Related Work}
In general, the related work could be roughly grouped into three parts, namely \textit{Social Relation Theory}, \textit{Social Relation Recognition in Computer Vision~(CV) area} and \textit{Transfer CLIP Knowledge into Multi-modal Tasks}.

\subsection{Social Relation Theory}
The study of social relationships~\cite{fiske1992four, duncan1981structure, tajfel1982social, heider2013psychology, de1960learning, abelson1968theories, de1958perception, de1959subjective, de1965social, de1962cognitive} is fundamental to social sciences, serving as the social cognitive sources across numerous domains from social-affect health research~\cite{tay2013social, holt2010social, umberson2010social} to socially interactive robots~\cite{de1960learning, beckers2000fom, deneubourg1991dynamics, fong2003survey}, as well as the visual social relation recognition fields~\cite{li2020visual, li2017dual, li2020graph, liu2019social, wu2021linking}.

A widely endorsed unified theory of social relations~\cite{fiske1992four} simplifies social relations into four elementary prototypes, i.e., communal sharing (CS), equality matching (EM), authority ranking (AR), and market pricing (MP). This theory posits that the motivation, planning, production, comprehension, coordination, and evaluation of human social life are largely based on combinations of these four psychological models. Different taxonomies~\cite{li2017dual, sun2017domain, liu2019social, vicol2018moviegraphs} of social relations can also be viewed as combinations of these four elementary forms of sociality. Each social relation exhibits unique properties~\cite{de1960learning, de1962cognitive, de1958perception, de1965social, abelson1968theories}; for example, CS and AR relations involve principles of reflexivity and transitivity, while the CS relation is symmetric, AR is antisymmetric~\cite{fiske1992four}. To model these characteristics and adopt a holistic group perspective, ConSoR introduces interpersonal modules, effectively representing the structure and patterns of the interpersonal behavior cycle~\cite{heider2013psychology, de1959subjective, de1958perception, tabbat1976anatomy, kerlinger1951decision}.

Further, social relations can be visually captured from a social-cognitive perspective. For instance, individuals engaged in Communal Sharing (CS) often seek similarity with others, a tendency that is particularly pronounced during adolescence. In this stage, conforming to the norms of one's reference group or clique in aspects like dress, speech, and preferences becomes crucial~\cite{heider2013psychology, fiske1992four}. Additionally, specific objects serve as static markers representing specific social relations. For example, European thrones, crowns, and scepters may symbolize Authority Ranking (AR) relations, while relics, roses, and wedding rings often represent CS relations. To model these nuances, we first employ a multi-modal side adapter tuning mechanism, to incorporate social-aware semantic knowledge from pre-trained CLIP models, enabling the model to understand what specific objects and interactions represent. Subsequently, we encourage the model to focus on these decisive social-visual context factors and common social patterns through a visual-linguistic contrasting approach, with social-aware descriptive language prompts.

\subsection{Social Relation Recognition in CV}
Social relation recognition in computer vision~(CV)~\cite{li2017dual,wang_deep_2018,li2020graph, wu2021linking, xu2021socializing, li2020visual, zhang2019multi, dai2019two, li2021hf, goel2019end, zhang2015learning, lin2022social, yang2021gaze, qing2021srr, wu2021linking, PengHXXZC23, qin2023fall} has attracted wide attention in both academia and industry recently. This task involves predicting the social relation between pairs of people in images, using both explicit and implicit cues within each individual image. Along this line, two large-scale datasets, namely The People in Photo Albums~(PIPA)~\cite{zhang_beyond_2015, sun2017domain} and the People in Social Context~(PISC)~\cite{li2017dual} are regarded as benchmarks for this task.

Most of the prior arts attempted to capture the potential information for social relation inference via explicit modeling of detected human traits~\cite{sun2017domain, zhang_learning_2015, zhang_beyond_2015} and objects~\cite{li2017dual}, or by incorporating knowledge graph~\cite{wang_deep_2018}. For instance, \cite{zhang_learning_2015} utilized a deep neural network to capture social relation-related attributes, such as gender, expression, and head pose, via detected human face. \cite{li2017dual} proposed a dual glance model where the first glance fixates on modeling the individual pair of interests and the second glance deploys the attention mechanism on context cues using detected objects. \cite{wang_deep_2018} exploited the benefits by integrating structured knowledge graphs constructed by the additional object detectors. Graph-based methods~\cite{goel2019end, li2020graph, wu2021linking, qin2023fall} attempt to model multiple relationships as graphs, exploring logical constraints among various social relations.
Though these models attempt to learn the specific social patterns based on human interactions or co-occurrence, they may ignore the critical yet subtle contextual social clues such as undetected objects due to the constraints of pre-trained object detectors. Additionally, they can hardly capture the intrinsic semantic connections among massive details due to the relatively small dataset. To this end, we obtain the semantic contextual information in a free-form way via the top-down attention mechanism, and further explore the social-aware semantics via the visual-linguistic contrasting approach.

\subsection{Transfer CLIP Knowledge into Multi-modal Tasks}
Pre-trained on millions of text-image pairs using a contrastive approach, CLIP~\cite{radford2021learning} exhibits a remarkable ability to align visual and language modalities, making it suitable for numerous multimodal downstream tasks~\cite{GuLKC22, xu2023side, fang2021clip2video, portillo2021straightforward, luo2022clip4clip, ramesh2022hierarchical, song2022clip}.
For instance, ViLD~\cite{GuLKC22} distills the knowledge from the teacher CLIP model into a two-stage student detector, while SAN~\cite{xu2023side} attaches a side network to reuse CLIP features and makes the predicted mask proposals semantically aware in the context of the semantic segmentation task. PointCLIP~\cite{zhang2022pointclip} designs an inter-view adapter to transfer 2D CLIP knowledge to the 3D recognition task. Inspired by them, ConSoR first utilizes a multi-modal side adapter network (MSAT) to transfer visual-linguistic semantics from CLIP, which can also mitigate the severe overfitting issues in training. We also find that the parameter-shared among both visual-linguistic adapters performs better since it establishes a more semantically aligned cross-modal space.
Besides, there are several previous works that attempt to explore the semantics of label text. Instead of mapping labels to numbers or one-hot vectors, ActionCLIP~\cite{wang2021actionclip} treats the action recognition task as a video-label text matching problem. DetCLIP~\cite{yao2022detclip} and CapDet~\cite{long2023capdet} further approach the recognition task as a region-category matching task and utilize language to unify different tasks. In contrast to merely using label text as language input, ConSoR constructs social-aware descriptive language prompts for images, and further encourages the models to focus on the social-decisive visual contexts through the visual-linguistic contrasting.

%% file: chapters/03_method.tex
\section{Methodology}

\begin{figure*}[ht]
  \includegraphics[width=1.0\textwidth]{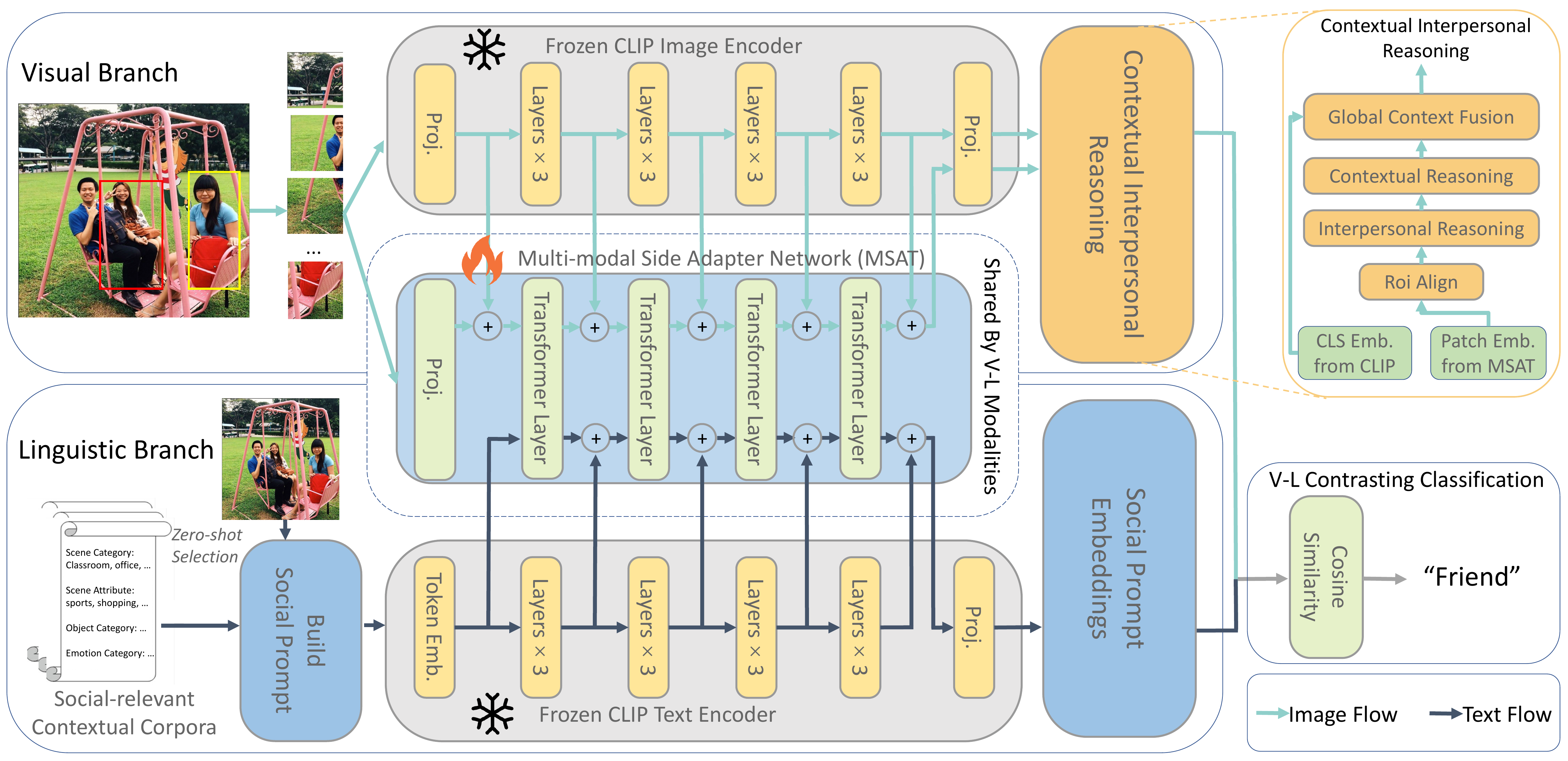}
  \vspace{-3mm}
  \caption{This figure depicts the ConSoR framework, where the frozen visual and linguistic CLIP encoders employ a shared multi-modal side adapter to learn social-aware representations. Then, the contextual interpersonal reasoning module extracts individual-pair features, followed by a CLIP-style visual-linguistic contrasting head for social relationship classification.}
  \label{fig:framework}
\end{figure*}

To address the aforementioned problems, we propose the ConSoR framework as shown in Figure~\ref{fig:framework}, which recognizes social relationships visually from a social cognitive perspective.
Specifically, given an image $I$ and a social-relation class set $R = \{r_1, r_2, \dots, r_C\}$ (where $C$ is the number of social relations), we extend the side adapter tuning approach~\cite{zhang2020side, Sung2022LST, xu2023side} to the multi-modal field, to transfer rich social-aware semantics from CLIP into our lightweight backbone.
Moreover, we perform relation reasoning from both contextual and interpersonal perspectives to examine the visual factors and common social patterns that may suggest social relations. 
Then, we construct social-aware descriptive language prompts for each image, and then encourage the model to focus on social-decisive contexts via a visual-linguistic contrasting approach.
We now turn to illustrate the technical details of the ConSoR framework in the following part.

\subsection{Multi-modal Side Adapter Tuning}
Specific objects and interactions such as crowns, roses, wedding rings and kiss often act as semantic markers of certain social relations. While these meanings are readily understood by humans, models, particularly when trained on relatively small datasets like those in the social relation recognition field, struggle to grasp these social concepts. Thus, in this section, we introduce the proposed Multimodal Side Adapter Tuning~(MSAT) module, which can efficiently transfer the social semantics from CLIP~\cite{radford2021learning} models into our lightweight multi-modal transformer backbone. By sharing parameters and latent knowledge across the visual and linguistic branches, MSAT not only obtains more semantically aligned representations across modalities, but also mitigated the severe overfitting issues compared with finetuning CLIP directly.

We denote the feature of the CLIP image and text encoder's $i$-th layer as $\mathcal{V}^{CLIP}_i\in \mathbb{R}^{L^v\times r^v}$ and $\mathcal{T}^{CLIP}_i\in \mathbb{R}^{L^t\times r^t}$, where $L^v$ and $L^t$ means the number of image patches and words, $r^v$ and $r^t$ is the visual and text feature dimensions.
For global image and text input embedding in CLIP, we denote the [CLS] token in image patches and [EOT] token in text sequences as $\mathcal{V}^{CLIP}\in \mathbb{R}^{r}$ and $\mathcal{T}^{CLIP}\in \mathbb{R}^{r}$~(where $r$ is the default dimension of CLIP).
In addition, $\mathcal{V}^{SN}_i$ and $\mathcal{T}^{SN}_i$ represent the image and text output of our multi-modal side adapter network's $i$-th layer.

As shown in the side adapter network part of Figure~\ref{fig:framework}, the input image is split into $16\times 16$ patches and then embed into patch embeddings via a linear embedding layer. Further, the position embeddings are added to form the image representation $\mathcal{V}^{SN}_0$ for the side adapter network. 
For text inputs, following~\cite{Sung2022LST}, we use the downsampled linear projections of the summation of word embeddings and position embeddings from the CLIP text encoder $\mathcal{T}^{CLIP}_0$ as the text inputs $\mathcal{T}^{SN}_0$ of the side adapter network, which significantly saves memory usage for a large number of word embeddings.

To efficiently transfer the rich social-aware semantics from CLIP into our lightweight backbone, we construct our lightweight side adapter network comprising just 4 transformer layers and it is worth noting that we are the first to explore the side adapter tuning in the multi-modal settings, finding that the parameter-shared MSAT achieves the best performance when tuning both modalities.
For the visual branch, we incorporate the \{0, 3, 6, 9, 12\}-th intermediate features of CLIP's image encoder into the \{0, 1, 2, 3, 4\}-th hidden states of the side adapter network, respectively. Similarly, for the linguistic branch, the \{3, 6, 9, 12\}-th intermediate features of CLIP's text encoder are integrated into the \{1, 2, 3, 4\}-th hidden states of the side adapter network.
Specifically, we fuse the $i$-th intermediate CLIP visual feature into the $j$-th layer of our side adapter network via the gated mechanism as follows:
\begin{equation}
    \hat{\mathcal{V}}^{SN}_j = \mu_j^v \mathcal{V}^{SN}_j + (1-\mu_j^v) \mathcal{V}^{CLIP}_i,
\end{equation}
\begin{equation}
    \mu_j^v = sigmoid(\frac{\alpha_j^v}{\tau}),
\end{equation}
where $\alpha_j^v$ is a learnable gate scalar for the visual branch and $\tau$ is temperature. 
Similarly, we fuse the latent linguistic knowledge from CLIP into our side adapter network. 
\begin{equation}
    \hat{\mathcal{T}}^{SN}_j = \mu_j^t \mathcal{T}^{SN}_j + (1-\mu_j^t) \mathcal{T}^{CLIP}_i,
\end{equation}
\begin{equation}
    \mu_j^t = sigmoid(\frac{\alpha_j^t}{\tau}),
\end{equation}
where $\alpha_j^t$ is a learnable gate scalar for the linguistic branch.

After fusing the multi-modal semantics from CLIP, we then utilize the side adapter network, which consists of multi-head self-attention (\textit{MSA}) layers~\cite{vaswani2017attention}, to obtain the representation of the image $\mathcal{V}^{SN}\in \mathbb{R}^{L^v\times r}$ and the feature of the corresponding social prompt $\mathcal{T}^{SN}\in \mathbb{R}^{L^t\times r}$ as follows,
\begin{equation}
    \mathcal{V}^{SN}_{j+1} = MSA(\hat{\mathcal{V}}^{SN}_j)
\end{equation}
\begin{equation}
    \mathcal{T}^{SN}_{j+1} = MSA(\hat{\mathcal{T}}^{SN}_j)
\end{equation}
By sharing parameters between the visual and linguistic spaces, the multi-modal side adapter network learns representations that are more semantically aligned across modalities, which significantly enhances the performance on this task.

\begin{figure}[h]
  \includegraphics[width=0.5\textwidth]{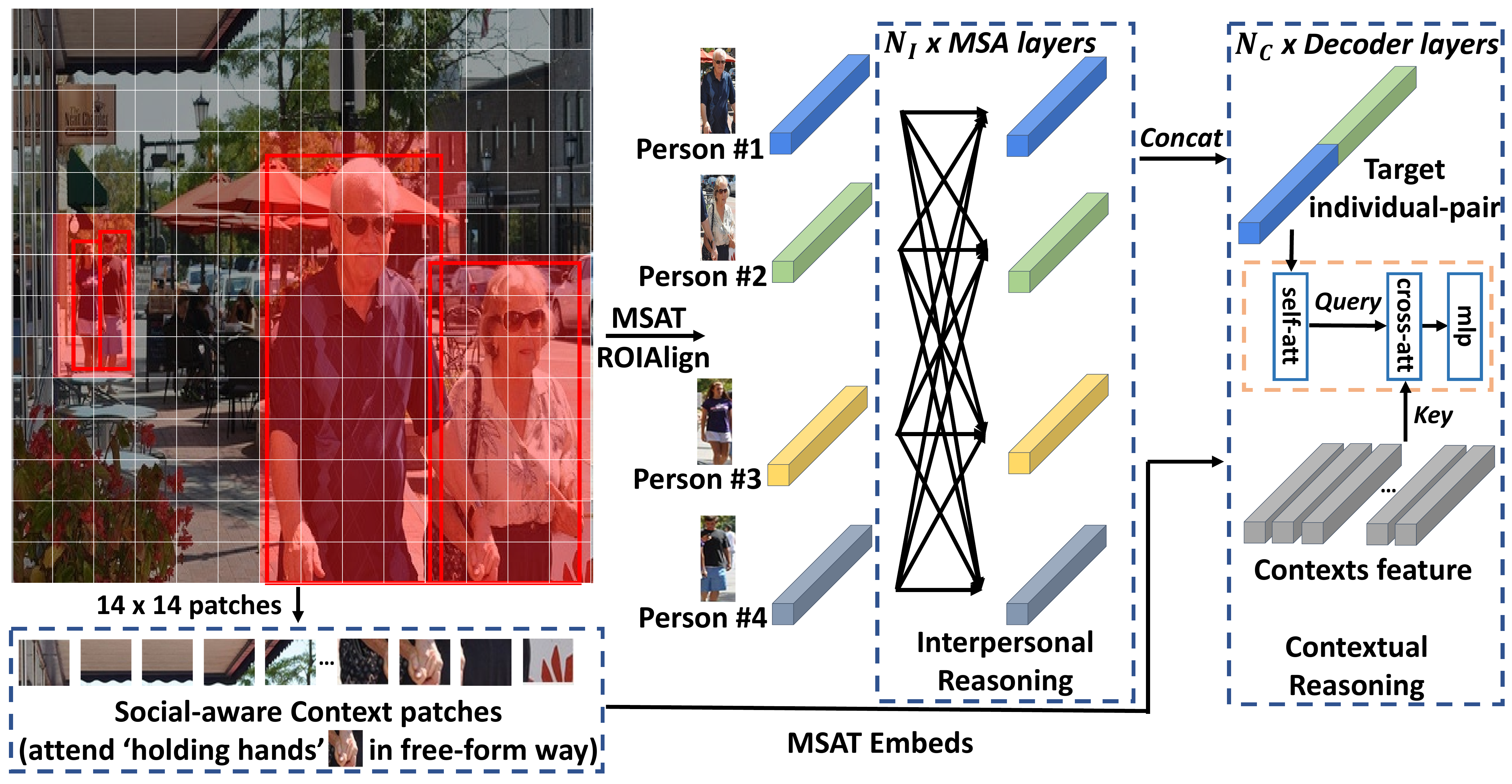}
  \vspace{-3mm}
  \caption{Illustration of our proposed interpersonal and contextual reasoning modules. The interpersonal reasoning module can model the crucial properties of social relations, such as transitivity and reflexivity. Furthermore, the model attends to social-aware contexts, such as `holding hands' in the image, through the contextual reasoning module.}
  \label{fig:cir_illustration}
\end{figure}

\subsection{Contextual Interpersonal Reasoning}
After obtaining the social-aware semantic knowledge through the MSAT module, we now turn our attention to examining which visual factors are beneficial for social relation inference. Taking into account the impact from interpersonal and contextual perspectives, we propose the Contextual Interpersonal Reasoning Module~(CIR), which consists of three parts, i.e., interpersonal reasoning, contextual reasoning, and global context fusion.
Some prior studies~\cite{li2017dual, sun2017domain} treat each individual pair as mutually independent, neglecting the influence of others. Although this approach can differentiate between individual relationships, it often falls short in performance due to the oversight of critical social relation properties such as transitivity, reflexivity, or symmetry~\cite{fiske1992four, de1960learning, de1962cognitive, de1958perception, de1965social, abelson1968theories}. Consider Figure~\ref{fig:intro} as an example: identifying the \textit{parent-child} relationship between the baby and the two adults can reinforce our understanding of the \textit{couple} relationship between the man and woman, demonstrating the principle of \textit{transitivity}. 

Therefore, as shown in Figure~\ref{fig:cir_illustration}, we utilize the interpersonal reasoning submodule to model this crucial social pattern. Given $N$ person bounding boxes $\{b_1, b_2, \dots, b_N\}$, we use \textit{ROIAlign}~\cite{he2017mask} operation to extract the features of the certain individual from the MSAT knowledge-injected visual feature $\mathcal{V}^{SN}$ as follows:

\begin{equation}
    \mathbf{P}_i = ROIAlign(\mathcal{V}^{SN}, b_i),
\end{equation}
where $\mathbf{P}_i\in \mathbb{R}^r$ is the person feature in the bounding box $b_i$. 

Then, we perform the interpersonal reasoning via the $N_I$ layers of multi-head self-attention (\textit{MSA}) mechanism,
\begin{equation}
    \hat{\mathbf{P}} = MSA(\mathbf{P})
\end{equation}
where $\mathbf{P} = \{\mathbf{P}_1, \dots, \mathbf{P}_N\}$ represents the concatenation of all $N$ individuals' features, and $\hat{\mathbf{P}} = \{\hat{\mathbf{P}}_1, \dots, \hat{\mathbf{P}}_N\}$ denotes the features incorporating interpersonal influences.

\begin{figure*}[t]
  \includegraphics[width=1.0\textwidth]{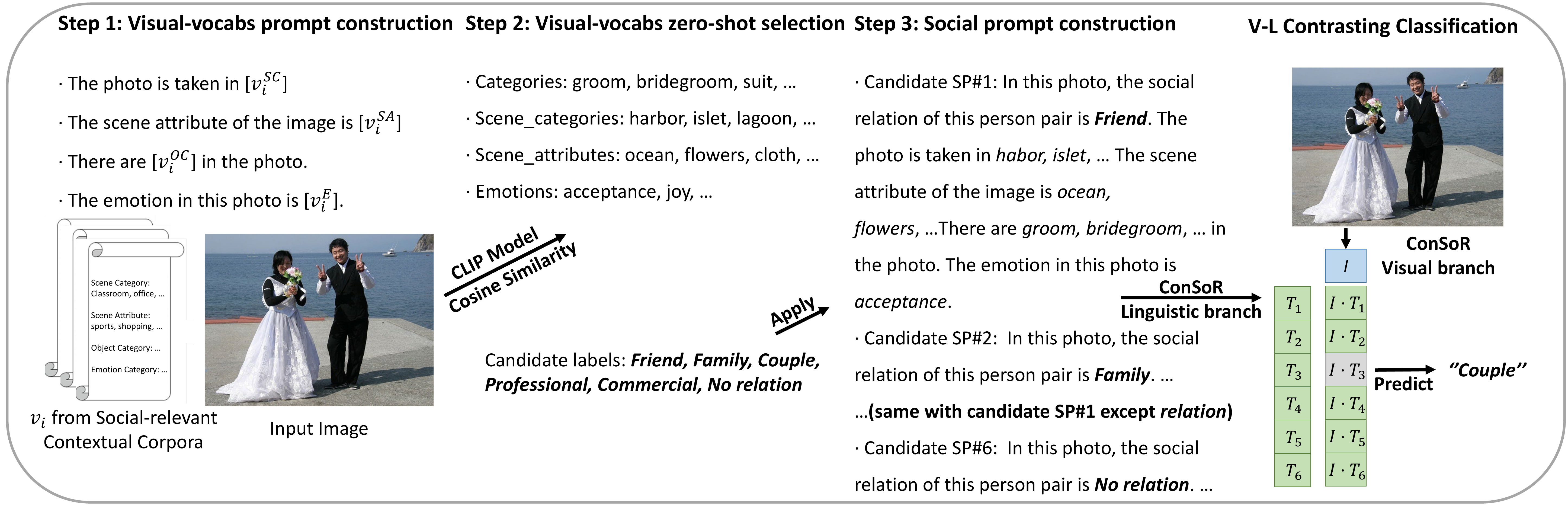}
  \caption{The pipeline for constructing descriptive contextual social prompts. With social prompts, ConSoR integrates rich linguistic information into visual social relation tasks without requiring additional annotations. This integration further enables the model to identify social-decisive contexts through the following visual-linguistic contrasting classification.}
  \label{fig:social_prompt}
\end{figure*}

Moreover, contextual information within an image plays a crucial role in inferring underlying social relations. For example, in an office setting characterized by laptops, monitors, and pens, the relationship between individuals is likely to be that of \textit{colleagues}. However, the focus on specific contextual elements should vary among different individuals. Consider a scene in a church crowded with people: when inferring the relationship between the bride and groom on stage, the model should concentrate on items like roses, the white dress, wedding rings, and their intimate interactions. In contrast, when judging the relationships of others in the same scene, the model should consider different contextual elements. 
Unfortunately, previous methods~\cite{zhang_learning_2015, zhang_beyond_2015, li2017dual, wang_deep_2018, liu2019social, wu2021linking} overlook this crucial social principle, they rely on additional object detectors~\cite{ren2015faster} to identify all the objects or pre-defined traits in the image and then feed all the detected contexts with the target individual-pair to infer the social relation. This kind of approach has three primary drawbacks: first, it requires heavy object detectors and significantly depends on their performance, which might miss subtle yet critical objects such as wedding rings, as well as lack of capturing interaction details like the \textit{holding hands} in Figure~\ref{fig:cir_illustration}; second, it applies the same contextual analysis to all individuals in the image, potentially introducing noise in the inference of specific individual-pair relations.

To this end, we propose the contextual reasoning submodule to discover the social-decisive visual cues that may suggest social relationships among individuals in a free-form manner, which can adaptively focus on specific social-decisive contexts for different individual-pairs. Specifically, we utilize a $N_C$ layers of Transformer-based Decoder to perform contextual reasoning given the global feature map $\mathcal{V}^{SN}$,
\begin{equation}
    \hat{\mathbf{U}} = Concat(\hat{\mathbf{P}_i}, \hat{\mathbf{P}_j})
\end{equation}

\begin{equation}
    \overline{\mathbf{U}} = Decoder(\hat{\mathbf{U}}, \mathcal{V}^{SN})
    \label{decoder}
\end{equation}
where $(\hat{\mathbf{P}_i}, \hat{\mathbf{P}_j})$ are the target individual-pair features. Through the cross-attention mechanism in Equation~\ref{decoder}, the potentially related objects could be discovered as shown in Figure~\ref{fig:visual}.

Further, to incorporate the global context information $\mathcal{V}^{CLIP}$ from CLIP visual branch into the pair feature $\overline{\mathbf{U}}$, we propose the global context fusion as follows,
\begin{equation} 
    \mathbf{z}=sigmoid(\mathbf{W}_g^{u}\overline{\mathbf{U}}+\mathbf{W}_g^{clip}\mathcal{V}^{CLIP}+\mathbf{b}_g),
\end{equation}
\begin{equation}
    \mathbf{U}=\mathbf{z}\circ\overline{\mathbf{U}} +(1-\mathbf{z})\circ\mathcal{V}^{CLIP},
\end{equation}
where $\mathbf{W}_g^{u}$ and $\mathbf{W}_g^{clip}$ are gated matrices for $\overline{\mathbf{U}}$ and $\mathcal{V}^{CLIP}$, $\circ$ means the element-wise multiplication and $\mathbf{b}_g$ for the bias term.
We finally obtain the feature of the given individual-pair fusing with comprehensive semantic contextual information $\mathbf{U}\in \mathbb{R}^{r}$ through the MSAT and CIR modules.

\subsection{Descriptive Social Prompt Construction via Self-generated Visual-vocabs}

Beyond simple one-hot vector classification, previous approaches~\cite{wang2021actionclip, yao2022detclip, long2023capdet} in CLIP downstream tasks have explored the use of semantic label text. However, these methods fall short as they do not fully leverage the rich contextual information available in the language, resulting in a lack of an intermediate understanding of the rationale behind category selection. To overcome this limitation, we introduce social-aware descriptive prompts (termed `Social Prompts') that articulate the contextual social states of images. This approach not only encourages models to focus on social-decisive factors through a visual-linguistic contrasting method, but also provides a degree of inherent explainability.
As illustrated in Figure~\ref{fig:social_prompt}, there are three steps to construct the descriptive social prompt: i). visual-vocabs prompt construction, ii). visual-vocabs zero-shot selection, and iii). social prompt construction.

In visual-vocabs prompt construction, we use $4$ social-relevant corpora to generate visual-vocabs based on the image, including scene categories $\{v_1^{SC},...,v_{|SC|}^{SC}\}$ and scene attributes $\{v_1^{SA},...,v_{|SA|}^{SA}\}$ from Place-365 dataset~\cite{zhou2017places}, objects categories $\{v_1^{OC},...,v_{|OC|}^{OC}\}$ from ImageNet-1k dataset~\cite{deng2009imagenet}, and emotion categories $\{v_1^{E},...,v_{|E|}^{E}\}$ from Plutchnik's Wheel of Emotions~\cite{mohsin2019summarizing}, where $|SC|$, $|SA|$, $|OC|$ and $|E|$ are the size of respective corpora. Each corpus is a distinct set of social-relevant domain-specific vocabulary and the statistic information and examples of corpora are shown in Table~\ref{tab:corpora}. 

Utilizing CLIP's robust zero-shot capabilities, we then select visual-vocabs from each corpus in a zero-shot manner. We construct prompts with templates depicted in Table~\ref{tab:template} for each vocab in the corpus. Given an image $I$ and all candidate visual-vocab prompts, we utilize CLIP dual encoder to encode and further filter out the related visual-vocabs $Vocab_I$ based on cosine similarity. We select top $5$ relevant visual-vocabs $Vocab_I=\{v_{t_1}^{SC},...,v_{t_5}^{SC},v_{t_1}^{SA},...,v_{t_5}^{SA},v_{t_1}^{OC},...,v_{t_5}^{OC},v_{t_1}^{E}\}$ for each corpus~(top $1$ for emotion corpus due to its relatively smaller size).

Further, we construct social prompts with self-selected $Vocab_I$. For each social relation $r_i \in R$, we prepend class name prompt \textit{"In this photo, the social relation of this person pair is [$r_i$]."} with visual-vocab prompt \textit{"The photo is taken in [$v_{t_1}^{SC}$],...,[$v_{t_5}^{SC}$]. This scene attribute of the image are [$v_{t_1}^{SA}$],...,[$v_{t_5}^{SA}$]. There are [$v_{t_1}^{OC}$],...,[$v_{t_5}^{OC}$] in the photo. This emotion in this photo is [$v_{t_1}^{E}$]."} to form the social prompt $P_I^{r_i}$. Then we encode $P_I^{r_i}$ with our side adapter backbone and use the [EOT] token from $\mathcal{T}^{SN}$ as the social prompt embedding, denoted as $\mathbf{T}^{sp}_{r_i}$ for the candidate social relation $r_i \in R$.

\begin{table}[!h]
    \caption{Social-Relevant Corpora}
    \setlength\tabcolsep{0.5mm}
    \begin{tabular}{lll}
    \hline
    Corpora & \#words & Examples \\ \hline
    Scene Category~\cite{zhou2017places} & $365$ & movie theater, classroom, office \\
    Scene Attribute~\cite{zhou2017places} & $94$ & sports, shopping, playing, camping\\
    Object Category~\cite{deng2009imagenet} & $1000$ & bow-tie, rubber, shopping basket \\
    Emotion Category~\cite{mohsin2019summarizing} & $24$ & joy, trust, interest, grief, surprise\\
    \hline
    \label{tab:corpora}
    \end{tabular}
\end{table}

\begin{table}[!h]
    \caption{Visual-vocabs prompt templates for each social-relevant contextual corpus.}
    \setlength\tabcolsep{0.5mm}
    \renewcommand{\arraystretch}{1.3}
    \begin{tabular}{ll}
    \hline
    Corpora & Prompt Template \\ \hline
    Scene Category~\cite{zhou2017places} & \textit{"The photo is taken in [$v_i^{SC}$]."} \\
    Scene Attribute~\cite{zhou2017places} & \textit{"The scene attribute of the image is [$v_i^{SA}$]."} \\
    Object Category~\cite{deng2009imagenet} & \textit{"There are [$v_i^{OC}$] in the photo."} \\
    Emotion Category~\cite{mohsin2019summarizing} & \textit{"The emotion in this photo is [$v_i^{E}$]."} \\
    \hline
    \label{tab:template}
    \end{tabular}
\end{table}

\begin{table*}[!t]
    \renewcommand{\arraystretch}{1.4}
    \setlength\tabcolsep{1.5mm}
    \caption{Experimental results on PISC and PIPA dataset. We outperform SOTA methods by achieving \textbf{+12.2 mAP} improvement on the PISC-Fine dataset, which is considered the primary metric for evaluating visual social relation recognition task. (Int.: Intimate, Non.: Non-Intimate, NoR.: No Relation, Fri.: Friend, Fam.: Family, Cou.: Couple, Pro.: Professional, Com.: Commercial.)
    The PIPA-Coarse/PIPA-Fine dataset is 5/16-class classification, respectively.}
    \label{pisc pipa}
    \centering
    \begin{tabular}{c|ccc|c|cccccc|c|c|c}
        \hline
        \multirow{2}*{Methods} & \multicolumn{4}{c|}{PISC Coarse} & \multicolumn{7}{c|}{PISC Fine} & PIPA Coarse & PIPA Fine \\
        \cline{2-14} 
        & Int. & Non. & NoR. & \textbf{mAP} & Fri. & Fam. & Cou. & Pro. & Com. & NoR. & \textbf{mAP} & \textbf{Acc@1} & \textbf{Acc@1} \\
        \hline
        Pair-Box-CNN~\cite{li2017dual} & 70.3 & 80.5 & 38.8 & 65.1 & 30.2 & 59.1 & 69.4 & 57.5 & 41.9 & 34.2 & 48.2 & 65.9 & 58.0 \\
        \cline{1-1}
        Dual-Glance~\cite{li2017dual} & 73.1 & 84.2 & 59.6 & 79.7 & 35.4 & 68.1 & 76.3 & 70.3 & 57.6 & 60.9 & 63.2 & - & 59.6 \\
        \cline{1-1}
        GRM~\cite{wang_deep_2018} & 81.7 & 73.4 & 65.5 & 82.8 & 59.6 & 64.4 & 58.6 & 76.6 & 39.5 & 67.7 & 68.7 & - & 62.3 \\
        \cline{1-1}
        MGR~\cite{zhang2019multi} & - & - & - & - & 64.6 & 67.8 & 60.5 & 76.8 & 34.7 & 70.4 & 70.0 & - & 64.4 \\
        \cline{1-1}
        SRG-GN~\cite{goel2019end} & - & - & - & - & 25.2 & 80.0 & 100.0 & 78.4 & 83.3 & 62.5 & 71.6 & - & 53.6 \\
        \cline{1-1}
        GR$^2$N~\cite{li2020graph} & 81.6 & 74.3 & 70.8 & 83.1 & 60.8 & 65.9 & 84.8 & 73.0 & 51.7 & 70.4 & 72.7 & 72.3 & 64.3 \\
        \cline{1-1}
        HF-SRGR~\cite{li2021hf} & 89.1 & 87.0 & 75.5 & 84.6 & 82.2 & 39.4 & 33.2 & 60.0 & 47.7 & 71.8 & 73.3 & - & 65.9 \\
        \cline{1-1}
        GA-GCN~\cite{yang2021gaze} & 87.3 & 86.0 & 89.8 & 87.7 & 63.1 & 73.5 & 78.3 & 82.7 & 76.8 & 71.8 & 73.6 & - & 66.6 \\
        \cline{1-1}
        SRR-LGR~\cite{qing2021srr} & 89.6 & 84.6 & 78.5 & 84.8 & 83.9 & 52.4 & 35.9 & 64.0 & 54.0 & 63.6 & 73.0 & - & 66.1 \\
        \cline{1-1}
        PRM~\cite{lin2022social} & - & - & - & - & 56.5 & 70.8 & 72.3 & 78.5 & 81.4 & 68.3 & 73.2 & 73.2 & 65.6 \\
        \hline
        \textbf{ConSoR~(Ours)} & 91.3 & 78.7 & 71.0 & \textbf{90.2~\small{(+2.5)}} & 80.9 & 80.2 & 75.8 & 90.4 & 45.2 & 76.1 & \textbf{85.8~\small{(+12.2)}} & \textbf{82.4~\small{(+9.2)}} & \textbf{76.4~\small{(+9.8)}} \\
        \hline
    \end{tabular}
\end{table*}

\subsection{Visual-linguistic Contrasting Classification via Social Prompt} \label{contrasting}
To further facilitate the model's focus on social-decisive contexts, we employ visual-linguistic contrasting between $\mathbf{U}$ from CIR module and the candidate social prompt representations $\{\mathbf{T}^{sp}_{r_1}, \dots, \mathbf{T}^{sp}_{r_C}\}$.
It is worth noting that $\mathbf{T}^{sp}_{r_i}$ are slightly different, as the candidate social prompts only vary in the prepended class name prompt. As a result, social-relevant contextual clues can be further examined by applying contrastive learning between $\mathbf{U}$ and $\mathbf{T}^{sp}_{r_i} \in \{\mathbf{T}^{sp}_{r_1}, \dots, \mathbf{T}^{sp}_{r_C}\}$. Specifically, we compute the cosine similarity between the individual-pair feature and all candidate social prompts features. Then, we apply softmax activation to compute the cross-entropy loss as follows, 
The loss for ConSoR can be written as:
\begin{equation}
    \mathbf{z} = [sim(\mathbf{T}^{sp}_{r_1}, \mathbf{U}), \dots, sim(\mathbf{T}^{sp}_{r_C}, \mathbf{U})]
\end{equation}
\begin{equation}
    L_{ConSoR} = L_{CE}(softmax(\mathbf{z}), y_u)
\end{equation}
where $sim(a, b) = a^\top b /(\Vert a \Vert \Vert b \Vert)$, $y_u$ denotes the social relation label of the individual-pair, and the $L_{CE}$ is the cross entropy loss.
By employing a visual-linguistic contrasting classification paradigm, ConSoR is able to adaptively and thoroughly investigate the crucial, social-decisive contexts.

%% file: chapters/04_exp.tex
\section{Experiments}
In this section, we conduct detailed experiments to validate the effectiveness of ConSoR.

\subsection{Experimental Settings}

\textbf{Datasets.}
We utilize two benchmark datasets of social relation recognition for effectiveness validation, i.e. PIPA dataset~\cite{sun2017domain} and PISC dataset~\cite{li2017dual}. 
The PIPA dataset divides social life into 5 broad domains, and 16 finer social relations. Following~\cite{sun2017domain,li2020graph}, we adopt the overall accuracy across all categories for evaluation on PIPA dataset. 
The PISC dataset features a hierarchical structure of 3 general relationship categories (\textit{intimate}, \textit{non-intimate}, and \textit{no relation}) and 6 detailed categories (\textit{friend}, \textit{family}, \textit{couple}, \textit{professional}, \textit{commercial}, and \textit{no relation}).
We follow the standard train/validation/test split~\cite{li2017dual} for fair comparison.
For the PISC dataset, following ~\cite{li2017dual}, evaluation of methods is conducted using per-class recall and mean Average Precision (mAP) metrics.

\noindent \textbf{Implementation Details.} 
We leverage AdamW~\cite{loshchilov2017decoupled} as the optimizer.
We set the learning rate of all trainable parameters to $10^{-4}$, weight decay to $0.05$, training epoch to $6$, batch size to $32$\footnote{We utilize Distributed Data Parallel training on $2$ GeForce RTX 3090 Ti GPUs and the batch size of each gpu is $16$.}.
We utilize CosineAnnealing learning rate schedule to adjust the learning rate.
We adopt the pre-trained CLIP ViT-B/16~\cite{radford2021learning} as vision encoder and the input images are resized to $224\times 224$.
For MSAT, we set the transformer to $4$ layers, the number of heads to $6$, the dimension of transformer layers to $192$, the patch size to $16$, and temperature $\tau=0.1$ in MSAT.
We set the transformer layer number of interpersonal reasoning and contextual reasoning to $1$ by default and the number of heads to $8$.

\subsection{Experimental Results}

We compare our ConSoR with the following methods:
\begin{itemize}[leftmargin=*]
    \item \textbf{Pair-box-CNN~\cite{li2017dual}}: Extracts features from two given individuals using ROI pooling, then concatenates these features and applies a linear layer for classification.
    \item \textbf{Dual-Glance~\cite{li2017dual}}: Features are extracted by sending two cropped individual patches and their union region to CNNs. Additionally, the introduced `second glance' mechanism utilizes surrounding proposals as contextual information to refine the predictions.
    \item \textbf{GRM~\cite{wang_deep_2018}}: Constructs a knowledge graph comprising persons and objects within an image, and employs a gated graph neural network to identify social relations.
    \item \textbf{MGR~\cite{zhang2019multi}}: The model utilizes both the person-object graph and the pose graph of individuals to represent actions between people and objects, as well as interactions among pairs of people. Subsequently, it infers social relations using a graph convolutional network.
    \item \textbf{SRG-GN~\cite{goel2019end}}: Employs a GRU~\cite{chung2014empirical} to process two types of information: single-image data and relational attributes. It performs inference based on these inputs and uses the final output of this inference process to classify social relations.
    \item \textbf{GR$^2$N~\cite{li2020graph}}: Reasons out all relationships using multiple relation graphs, which incorporate logical constraints among various social relations.
    \item \textbf{HF-SRGR~\cite{li2021hf}}: Utilizes a social relation graph to extract and fuse hybrid features. This approach incorporates an attention mechanism into Graph Neural Networks (GNNs)~\cite{zhou2020graph} to generate features and interactions between pairs, thereby enhancing the reasoning of interactions.
    \item \textbf{GA-GCN~\cite{yang2021gaze}}: Introduces a gaze-aware graph convolutional network designed to facilitate context-aware social relation inference. This is achieved through a two-stream graph inference process that incorporates both gaze-aware attention and distance-aware attention mechanisms.
    \item \textbf{SRR-LGR~\cite{qing2021srr}}: Leverages graph neural networks to deduce social relations by analyzing interactions (local information) between social entities and the global contextual information embedded in the constructed scene-relation graph.
    \item \textbf{PRM~\cite{lin2022social}}: Two ResNet~\cite{he2016deep} modules are employed: one to extract character features, and the other to concentrate on the details of the entire graph, thereby constructing a graph for each relationship.    
\end{itemize}

The overall results are summarized in Table~\ref{pisc pipa}. 
Through incorporating rich semantics from CLIP models as well as encouraging models to focus on the crucial social-decisive contexts via the visual-linguistic approach, ConSoR significantly outperforms previous methods, achieving a \textbf{+12.2 mAP increase} on the PISC-Fine dataset, which is considered the acknowledged primary setting in the social relation recognition task. The improvement across most categories suggests that ConSoR exhibits greater robustness compared to other methods.

\begin{table*}[t]
    \renewcommand{\arraystretch}{1.4}
    \setlength\tabcolsep{2.9mm}
    \caption{Comparison with ConSoR Variants. We performed ablations five times and conducted T-tests between each ablation and ConSoR. The results indicate that ConSoR is statistically significantly superior in most cases.}
    \centering
    \begin{tabular}{l|c|c|c|c|c|c|c|c}
        \hline
        \multirow{3}*{Methods} & \multicolumn{4}{c|}{PISC~(mAP)}  & \multicolumn{4}{c}{PIPA~(Acc@1)} \\
        \cline{2-9}
        & \multicolumn{2}{c|}{Coarse} & \multicolumn{2}{c|}{Fine} & \multicolumn{2}{c|}{Coarse} & \multicolumn{2}{c}{Fine} \\
        \cline{2-9}
        & mean \& std. & p-value & mean \& std. & p-value & mean \& std. & p-value & mean \& std. & p-value \\
        \hline
        CLIP-ZeroShot & 36.20$\pm.00$ & $10^{-11}$ & 30.40$\pm.00$ & $10^{-9}$ & 34.80$\pm.00$ & $10^{-9}$ & 39.00$\pm.00$ & $10^{-7}$ \\
        CLIP-Frozen & 86.60$\pm.21$ & $10^{-9}$ & 77.72$\pm.61$ & $10^{-8}$ & 78.72$\pm.54$ & $10^{-5}$ & 73.36$\pm.43$ & 0.003 \\
        CLIP-Finetune & 88.22$\pm.26$ & $10^{-6}$ & 82.08$\pm.40$ & $10^{-6}$ & 77.82$\pm.67$ & $10^{-6}$ & 73.00$\pm.37$ & 0.002 \\
        ConSoR-Visual & 90.34$\pm.30$ & 0.109 & 84.62$\pm.45$ & 0.094 & 82.38$\pm.54$ & 0.877 & 75.84$\pm.36$ & 0.389 \\
        ConSoR-ObjDet & 89.94$\pm.25$ & 0.513 & 84.76$\pm.66$ & 0.274 & 81.28$\pm.57$ & 0.017 & 76.24$\pm1.21$ & 0.860 \\   
        \hline
        \textbf{ConSoR} & \textbf{90.24}$\pm.21$ & - & \textbf{85.80}$\pm.51$ & - & \textbf{82.44}$\pm.64$ & - & \textbf{76.38}$\pm1.22$ & - \\
        \hline
    \end{tabular}
    \label{self baseline}
\end{table*}

It is worth noting that Dual-Glance~\cite{li2017dual} employs an additional object detector~\cite{ren2015faster} to include contextual object information. It then concatenates these object features with the individual-pair feature for social relation prediction. However, this approach can impede the model's performance when the contextual objects are non-informative or potentially cause interference.
Though GRM~\cite{wang_deep_2018} and MGR~\cite{zhang2019multi} advance this approach by introducing a graph attention mechanism, which adaptively reasons about the most relevant contextual objects according to its proposed graph structure. Two significant drawbacks remain: 1) The performance of these models heavily relies on the accuracy of detection results. Crucial yet subtle social-decisive contexts, such as wedding rings, are essential for social relation judgment but often overlooked by these detector-based models. 2) The limited size of datasets in the visual social relation recognition field hinders the model's ability to learn the specific meanings of certain objects. For instance, understanding the relevance of thrones, crowns, and scepters for professional relations, or roses and wedding gowns for couple relations, remains a challenge.
However, our detector-free ConSoR overcomes these limitations by first learning rich, social-aware semantics from pre-trained CLIP models through the Multi-modal Side Adapter Network module. It then attends to social-decisive contexts in a free-form manner, enabled by the Contextual Interpersonal Reasoning module, which allows for a more nuanced comprehension of visual social relationships among individuals from a social cognitive perspective.

In addition, these previous models have overlooked the wealth of contextual information present in language, treating this task merely as a classic classification paradigm, i.e., mapping labels to one-hot vectors. In contrast, our ConSoR innovatively models this task as a multi-modal problem, requiring no additional annotations. It constructs social-aware descriptive language prompts for each image and then encourages the model to focus on social-decisive contexts through a visual-linguistic contrasting approach.


\begin{table}[h]
    \renewcommand{\arraystretch}{1.4}
    \setlength\tabcolsep{0.5mm}
    \caption{Efficiency analysis of ConSoR variants. All experiments were conducted using two Nvidia GeForce RTX 3090 Ti GPUs with 16 batch-size. ConSoR strikes a good balance between performance and complexity/efficiency}
    \centering
    \begin{tabular}{l|c|c|c}
        \hline
        Methods & Train.Params.(M) & TotalParams.(M) & GPUMem.(MB/GPU) \\
        \hline
        CLIP-ZeroShot & 0 & 149 & 1,628 \\
        CLIP-Frozen & 7.6 & 157 & 2,256 \\
        CLIP-Finetune & 157 & 157 & 5,706 \\
        ConSoR-Visual & 19.4 & 105 & 2,760 \\
        ConSoR-ObjDet & 20 & 214 & 4,535 \\
        \hline
        \textbf{ConSoR} & \textbf{20} & \textbf{170} & \textbf{3,554} \\
        \hline
    \end{tabular}
    \label{variants_efficiency}
\end{table}

\subsection{Ablation Study}
To verify the contribution of each module in our ConSoR framework, we further design several variants as well as ablations to conduct the ablation study on our main dataset~(i.e., PISC-Fine dataset). 

\subsubsection{Comparison with ConSoR variants.}
We examine the benefits of each module in ConSoR by comparing the following variants.
\begin{itemize}[leftmargin=*]
    \item \textbf{CLIP-ZeroShot}: Employ only the raw image to perform zero-shot inference in social relation recognition.
    \item \textbf{CLIP-Frozen}: Employ the \textit{ROIAlign} operation~\cite{he2017mask} to extract visual representations of individuals from features encoded by the fixed CLIP image encoder. Subsequently, uses a linear layer to classify the concatenated features of individual pairs.
    \item \textbf{CLIP-Finetune}: Similar to CLIP-Frozen, but finetune all CLIP parameters.
    \item \textbf{ConSoR-Visual}: Use a side adapter network to fine-tune the CLIP image encoder and utilizes the Contextual Interpersonal Reasoning (CIR) module for relation reasoning. Subsequently, it predicts social relations using a linear layer.
    \item \textbf{ConSoR-ObjDet}: Similar to ConSoR, we replace global feature $\mathcal{V}^{SN}$ in CIR decoder with detected-objects, following the same detector setting in~\cite{wang_deep_2018}. 
\end{itemize}

\begin{figure*}[!t]
	\centering
        
         \caption{Ablation for ConSoR on PISC-Fine dataset. In Figure~\ref{fig:ablation}a, ConR, IntR, GCF represents Context Reasoning, Interpersonal Reasoning and Global Context Fusion in CIR module. In Figure~\ref{fig:ablation}b, SC, SA, OC, E means scene category, scene attribute, object category and emotion corpora, respectively.}
	\subfloat{
		\includegraphics[width=0.95\textwidth, height=0.22\textwidth]{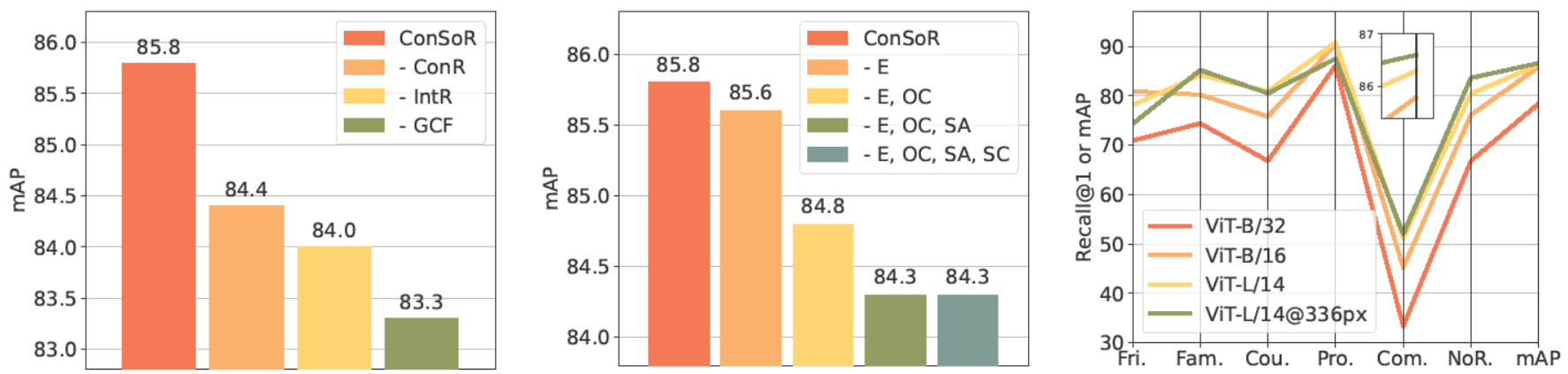}}
	\label{fig:ablation}
\end{figure*}

\begin{table*}[t!]
    \renewcommand{\arraystretch}{1.5}
    \setlength\tabcolsep{2.0mm}
    \caption{Ablations on MSAT Modules. The incorporation of MSAT effectively mitigates severe overfitting issues and enhances efficiency.}
    \centering
    \begin{tabular}{l|c|c|c|c|c|c}
        \hline
        \multirow{2}*{Methods} & \multicolumn{2}{c|}{PISC~(mAP)}  & \multicolumn{2}{c|}{PIPA~(Acc@1)} & Trainable Params. & GPU Memory\\
        \cline{2-5}
        & Coarse & Fine & Coarse & Fine & (Million) & (MB/GPU) \\
        \hline
        ConSoR-Visual w/o MSAT & 89.36$\pm.37$ & 83.58$\pm.42$ & 78.10$\pm.33$ & 72.98$\pm.70$ & 102 & 5,898 \\
        \textbf{ConSoR-V} & \textbf{90.34}$\pm.30$ & \textbf{84.62}$\pm.45$ & \textbf{82.38}$\pm.54$ & \textbf{75.84}$\pm.36$ & \textbf{19.4} & \textbf{2,760} \\
        \hline
        ConSoR w/o MSAT & 66.94$\pm2.72$ & 57.32$\pm3.53$ & 66.46$\pm1.66$ & 60.88$\pm2.97$ & 166 & 9,596 \\
        \textbf{ConSoR} & \textbf{90.24}$\pm.21$ & \textbf{85.80}$\pm.51$ & \textbf{82.44}$\pm.64$ & \textbf{76.38}$\pm1.22$ & \textbf{20} & \textbf{3,554} \\
        \hline
    \end{tabular}
    \label{msat_efficiency}
\end{table*}

As illustrated in Table~\ref{self baseline}, firstly, it can be observed that CLIP-ZeroShot notably underperforms compared to the other baselines and variants. This can be attributed to the significant domain gaps that CLIP's zero-shot capability encounters in the social relation recognition domain, which renders its direct application quite challenging.
Subsequently, we employ the ROIAlign operation to extract individual-pair features and utilize a single linear layer to bridge the domain gap. The improved performance of CLIP-Frozen, which surpasses previous methods~\cite{li2017dual,wang_deep_2018,li2020graph}, illustrates the great potential of integrating rich semantic knowledge into the field of social relation recognition.
We further observe that the model encounters severe overfitting issues with CLIP-Finetune, as the pre-trained model's size is not compatible with the relatively small dataset.
Additionally, we find that the incorporation of MSAT can effectively reduce the overfitting problem. Besides, the integration of the CIR module allows the model to capture contextual social clues, leading to a significantly improved performance in comparison to other baselines. As demonstrated in Table~\ref{variants_efficiency}, our novel multi-modal side adapter mechanism significantly reduces the model's complexity, making it much more lightweight (Trainable Parameters: $157M \rightarrow 19.4M$).
By incorporating language information from descriptive social prompts and further directing the model's focus towards social-decisive visual contexts through a visual-linguistic contrasting approach, ConSoR achieves superior performance compared to the ConSoR-Visual.
It is worth noting that the performance of ConSoR-Visual on the PISC-Coarse dataset is quite similar to ConSoR, as the task involves a relatively simple 3-class classification and we observe that the performance reaches a limit due to the presence of ambiguous relation annotations within the dataset.
Moreover, despite utilizing external heavy detectors, ConSoR-ObjDet exhibits inferior performance compared to our approach. This is because our simple yet effective CIR module is more adept at capturing implicit, decisive social context cues. As illustrated in Table~\ref{variants_efficiency}, ConSoR achieves an optimal balance between performance and computational complexity/efficiency.

\subsubsection{Impact of MSAT}
In this part, we validate the effectiveness of MSAT in ConSoR.
We propose four variants of MSAT:
\begin{itemize}[leftmargin=*]
    \item \textbf{w/o SN}: Utilizes a fixed CLIP encoder without a side adapter network.
    \item \textbf{TextSN/VisualSN}: Employs a single side adapter network for fine-tuning the CLIP text or visual encoders.
    \item \textbf{DualSN}: This variant incorporates two side adapter networks, attaching them separately to the CLIP visual encoder and the text encoder, with no shared parameters between the two branches.
    \item \textbf{with vanilla ViT/BERT}: Replaces the CLIP encoders with the standard vanilla versions of ViT~\cite{dosovitskiy2020image}~\footnote{https://huggingface.co/google/vit-base-patch16-384} and BERT~\cite{kenton2019bert}~\footnote{https://huggingface.co/google/bert\_uncased\_L-12\_H-512\_A-8}, each trained separately on unimodal datasets. The only difference from ConSoR lies in the encoder weights.
\end{itemize}

\begin{table}[H]
\caption{Performance of different side adapter network methods on PISC-Fine dataset.}
    \centering
    \renewcommand{\arraystretch}{1.0}
    \setlength{\tabcolsep}{5mm}{
    \begin{tabular}{l|l|c}
        \hline
        Methods & mAP & Trainable Params. (M) \\
        \hline
        w/o SN & 83.4 & 16.5 \\
        TextSN & 84.8 & 18.9 \\
        VisualSN & 84.5 & 19.4 \\
        DualSN & 84.5 & 21.8 \\
        with vanilla ViT/BERT & 76.2 & 20.0 \\
        MSAT & \textbf{85.8} & 20.0 \\
        \hline
    \end{tabular}}
    \label{msat}
\end{table}

\begin{table}[h]
    \setlength\tabcolsep{1.0mm}
    \renewcommand{\arraystretch}{1.3}
    \caption{Ablation study of different fusion strategies for MSAT on PISC-Fine dataset. Each fusion layer contains the text and visual features specific to its corresponding layer.}
    \label{tab:fusion}
    \centering
    \setlength{\tabcolsep}{6mm}{
    \begin{tabular}{c|c|c}
        \hline
        Strategies & Fusion Layers & mAP \\
        \hline
        w/o fusion & - & 79.8 \\
        \hline
        \multirow{5}*{Single-layer Fusion}& 0th layer & 80.2 \\
        & 3rd layer & 81.7 \\
        & 6th layer & 82.6 \\
        & 9th layer & 83.5 \\
        & 12th layer & 84.1 \\
        \hline
        \multirow{4}*{Multi-layers Fusion}& (9,12)-layers & 85.1 \\
        & (6,9,12)-layers & 85.5 \\
        & (3,6,9,12)-layers & 85.6 \\
        & (0,3,6,9,12)-layers & \textbf{85.8} \\
        \hline
    \end{tabular}}
    \label{fusion}
\end{table}

\begin{figure*}[!t]
	\centering
	\subfloat{
		\includegraphics[width=0.99\textwidth, height=0.636\textwidth]{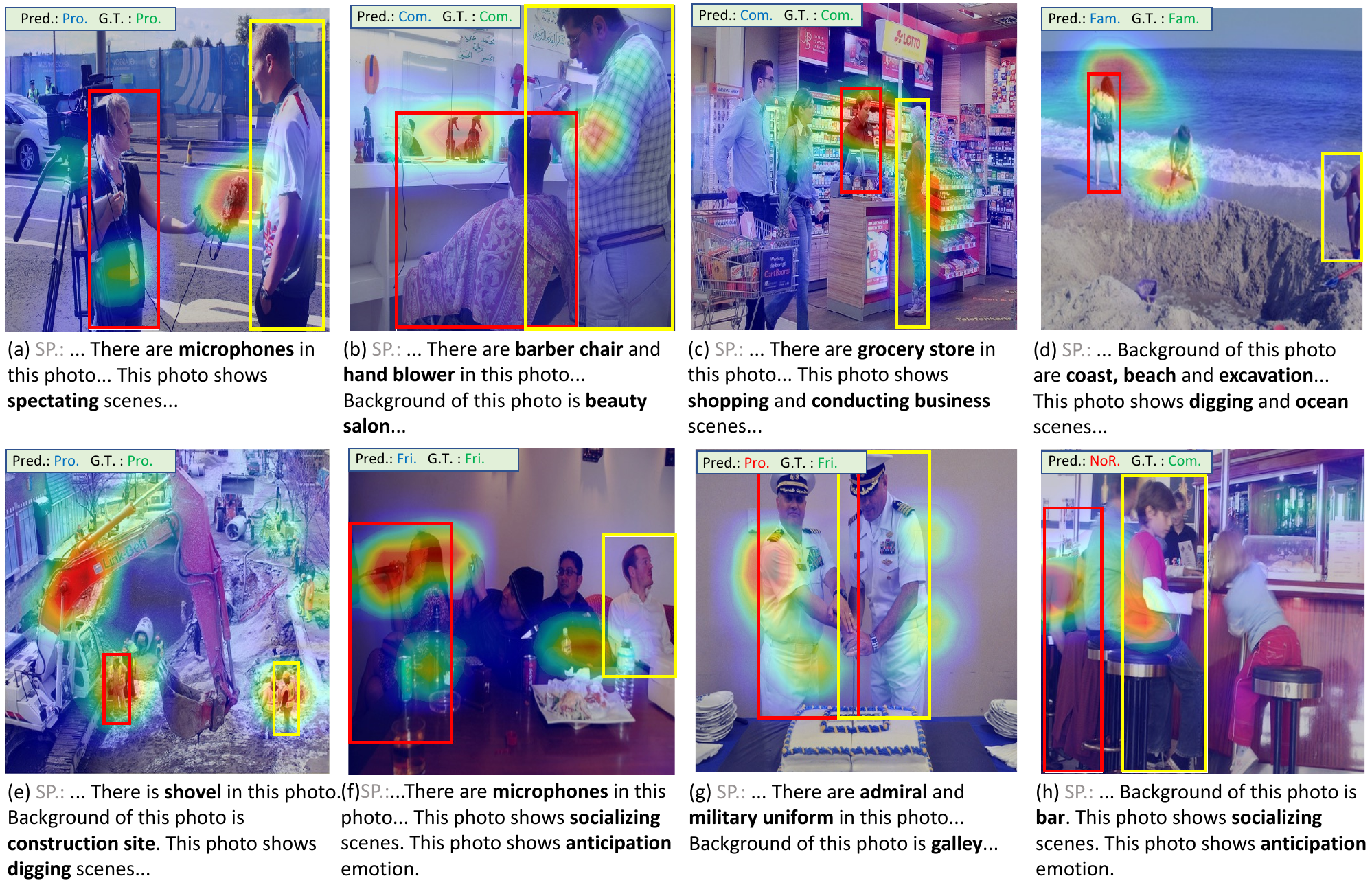}}
  
	\caption{The visualization cases of ConSoR on PISC-Fine dataset. Texts in green, blue, and red stand for ground truth, correct predictions, and wrong predictions, respectively. SP. means social prompt. Best viewed in color.}
	\label{fig:visual}
\end{figure*}

As presented in Table~\ref{msat}, in the absence of a side adapter network, the model struggles to adapt to the specific domain and exhibits poorer performance compared to the model equipped with the side adapter network. Additionally, we observe that multi-modal signals provide mutual benefits, as both TextSN and VisualSN underperform MSAT despite having similar numbers of trainable parameters. 
Moreover, though DualSN has more parameters than MSAT, by sharing parameters between the visual and linguistic branches, MSAT establishes a more semantically aligned cross-modal space, resulting in improved performance compared to DualSN.
Furthermore, compared to training on unimodal data, CLIP acquires rich semantics through contrastive learning on a massive scale of 400 million image-text pairs, a process crucial for understanding social relations in ConSoR. Consequently, using the standard vanilla versions of ViT~\cite{dosovitskiy2020image} and BERT~\cite{kenton2019bert}, which are trained solely on unimodal data, proves to be suboptimal ($85.8 mAP \rightarrow 76.2 mAP$) as semantic knowledge sources in ConSoR.

As demonstrated in Table~\ref{msat_efficiency}, the incorporation of MSAT not only enhances performance by mitigating severe overfitting concerns associated with directly applying the CLIP model as an encoder for relatively smaller datasets, but also significantly improves the model's efficiency.

To further examine the effectiveness of fusion strategies for MSAT, we conduct an ablation study of various fusion strategies, as shown in Table~\ref{tab:fusion}. We find that incorporating CLIP features can significantly boost performance, and integrating more layers leads to improved results. Additionally, the deeper the layer of information integrated into MSAT, the better its performance, as deeper features contain more high-level semantic information.



\subsubsection{Influence of CIR.}
As illustrated in Figure~\ref{fig:ablation}a, we observe that incorporating interpersonal and contextual reasoning significantly improves the model's performance. Through context reasoning, ConSoR more effectively identifies socially decisive contexts, such as wedding rings, flowers, or scepters. Moreover, the ConSoR decoder-based modeling approach not only achieves superior performance compared to detector-based methods but is also more lightweight.
Through interpersonal reasoning, ConSoR effectively models crucial social relation properties such as transitivity, reflexivity, and symmetry in a straightforward yet efficient manner. Additionally, the inclusion of global features significantly enhances performance, as the entire background contributes to the understanding of social relations.


\subsubsection{Influence of Descriptive Contextual Social Prompt.}
We conducted an ablation study on the visual-vocab corpora to assess the impact of descriptive contextual social prompts, as shown in Figure~\ref{fig:ablation}b. The findings reveal that a corpus enriched with social-relevant content enhances performance. This improvement is attributed to the increased semantic precision of social prompts describing the image, which more effectively guides the model to identify social-decisive cues through subsequent visual-linguistic contrasting.

\subsubsection{Impact of CLIP Backbones.}
The results presented in Figure~\ref{fig:ablation}c show that ConSoR's performance enhances as the backbone expands. A larger backbone, containing more semantic knowledge, underscores the excellent scalability of ConSoR.




\subsection{Case Study}
Finally, we turn to discuss some interesting rules based on several case studies, which are summarized in Figure~\ref{fig:visual}. 
Specifically, Figure~\ref{fig:visual}a and \ref{fig:visual}b demonstrate that by learning the social-aware semantic knowledge from CLIP, ConSoR can identify specific elements and contexts within the image that are essential for inferring social relations. 
For example, as shown in Figure~\ref{fig:visual}a, the model focuses on the microphone and camera, indicating that the situation is more likely a \textit{professional} interview context.

Moreover, we have noticed that the CIR module plays a crucial role in identifying social relationships. As shown in Figure~\ref{fig:visual}c, ConSoR demonstrates a greater focus on contextual information, such as items on the shelf, which can be attributed to the context reasoning submodule. 

Furthermore, as shown in Figure~\ref{fig:visual}d, ConSoR accurately infers the relationship between two distant individuals by detecting the presence of an intermediate girl. This demonstrates the `transitivity' in social relations, highlighting the significance of our interpersonal reasoning submodule.

Besides, Figure~\ref{fig:visual}e and \ref{fig:visual}f illustrate the benefits of integrating descriptive contextual social prompts into ConSoR, as the inclusion of \textit{construction site} and \textit{socializing} prompts improve the model's robustness on inferring social relations.

Finally, as shown in Figure~\ref{fig:visual}g and \ref{fig:visual}h, we observe that ConSoR cannot predict ``correctly" when faced with ambiguous labels, which is a challenging task even for humans. It is interesting that ConSoR does not attend to the \textit{holding hands} aspect in Figure~\ref{fig:visual}g, as it does not contribute to the social state.

%% file: chapters/05_conclusion.tex
\section{Conclusion}
In this paper, we introduce ConSoR, to recognize visual social relationships from a social cognitive perspective. By integrating social-aware semantics derived from CLIP models, ConSoR achieves a deeper understanding of social-aware contexts. This enhanced understanding enables it to excel at identifying critical visual contextual evidence, revealing social relations through a visual-linguistic contrasting approach.
Extensive experiments on two benchmark datasets demonstrate the superiority of ConSoR.
By accurately interpreting social relations from social cognitive perspectives, our work provides a solid foundation for a better understanding of human behavior and society.